\RequirePackage{fix-cm}
\documentclass[smallextended]{svjour3}       % onecolumn (second format)
\smartqed  % flush right qed marks, e.g. at end of proof
\usepackage{graphicx}
\usepackage{multirow}
\usepackage{rotating}
\usepackage{subfig}
\usepackage{times}
\usepackage{amsmath}
\usepackage{amssymb}
\usepackage{algorithm}
\usepackage{algorithmic}
\usepackage{array}
\usepackage{lineno, hyperref}
\usepackage{bm}
\usepackage{subfig}
\usepackage{enumerate}
\usepackage{float}

\begin{document}

\title{Deep Representation Learning for Road Detection using Siamese Network}


\author{Huafeng Liu \and Xiaofeng Han \and Xiangrui Li \and Yazhou Yao \and Pu Huang \and Zhenming Tang}


\institute{
Huafeng Liu, Xiaofeng Han, Xiangrui Li, Yazhou Yao, Zhenmin Tang \at School of Computer Science and Engineering, Nanjing University of Science and Technology, Nanjing 210094, China.\\
\email{liu.hua.feng@outlook.com}
\and
Pu Huang \at Jiangsu Key Laboratory of Big Data Security \& Intelligent Processing, Nanjing University of Posts and Telecommunications, Nanjing 210023, China.\\
}

\date{Received: date / Accepted: date}

\maketitle

\begin{abstract}
Robust road detection is a key challenge in safe autonomous driving. Recently, with the rapid development of 3D sensors, more and more researchers are trying to fuse information across different sensors to improve the performance of road detection. Although many successful works have been achieved in this field, methods for data fusion under deep learning framework is still an open problem. In this paper, we propose a Siamese deep neural network based on FCN-8s to detect road region. Our method uses data collected from a monocular color camera and a Velodyne-64 LiDAR sensor. We project the LiDAR point clouds onto the image plane to generate LiDAR images and feed them into one of the branches of the network. The RGB images are fed into another branch of our proposed network. The feature maps that these two branches extract in multiple scales are fused before each pooling layer, via padding additional fusion layers. Extensive experimental results on public dataset KITTI ROAD demonstrate the effectiveness of our proposed approach.

\keywords{Road Detection \and Siamese Network \and Data Fusion \and Deep Learning}
\end{abstract}

\section{Introduction}
\label{intro}

Road detection, also known as road segmentation, is a key challenge in autonomous land vehicle research during the last decade \cite{Xiaofeng2018,Luca2017,Charles2017,Chen2017}. It enables autonomous vehicles to navigate automatically in complex road conditions. The most widely applied solution is a monocular camera with road segmentation or boundary detection algorithms \cite{treml2016speeding,deepdrive,siam2017deep,teichmann2016multinet}. Though many fruitful works have been achieved, due to the projective property of camera, few metric and 3D spacial information can be recovered. Therefore, the limitation of those solutions is still obvious: they rely too much on well illumination and weather condition. Light Detection and Ranging (LiDAR) sensors are designed to strengthen the weakness mentioned above. LiDARs perceive by receiving reflected laser light, therefore, illumination condition is irrelevant. In addition, the 3D spacial metric information can be recovered easily by ToF at the same time.

Recently, deep learning is a powerful tool to process multimedia information \cite{fm2016,fm2017,Long2015Fully,yao2018tip,yao2018ijcai,YYZ2018,YYZ2017}, recomandation application\cite{YYZ2019,ijcai2019,Cheng2018,Cheng2018ijcai}, video understanding\cite{Nie2017,Song2018,tmm2018,aim,huangpu,mmm} and video analysis\cite{fm2018,HeKaiM,tkde2019,neurocom,ZhuLei2017,Xie2017,LiJ2018,ZhuL2017b,LiJ2018}. As a common task, road detection or segmentation on autonomous vehicles equipped with both camera and LiDAR is a active research area. On a autonomous vehicle, cameras and LiDARs can capture heterogeneous information simultaneously, thus making stable and accurate road detection possible. For camera sensors, its image acquisition is a consecutive sequence of perspective projection of environment. The camera imaging result is dense and informative in color and texture but lack of metric and spacial knowledge. As for LiDARs, the point cloud is usually obtained by scanning environment using laser range finders. It is easy to extract accurate metric and 3D spacial information from LiDAR results but point cloud is too sparse to support detection or segmentation tasks at a far distance. Methods using either of those two kinds of sensors have been widely studied in previous works\cite{Xiaofeng2018,Luca2017,Charles2017,Chen2017,prl,yao2016icme,XGS2019,acmmm}. However, to make autonomous vehicles smarter, researchers are trying to combine two sensors to draw on each other's strength. There are several works focusing on fusing the LiDAR and image information\cite{Schlosser,asvadi}, the most common idea is creating depth enhanced image by an up-sampling algorithm, then, low-level features are extracted to train different classifiers that can detect target.

Our paper is organized as follows: in section 2 we will briefly introduce the early and recent works on road detection and explain the Siamese deep learning method; in section 3, we present sparse LiDAR image generation as well as the proposed network architecture; in section 4, we discuss the experimental details and results of our methods on KITTI ROAD dataset; As a conclusion, we summarized this work and introduce our future works in section 5.

\section{Related Works}

Image or LiDAR based methods have been widely studied in early researches. Image based road detection solution is popular as it is low cost while informative in color and texture. Image road detection methods work in a pixel labeling way which classify the pixels or regions into road and non-road. Those researches look for texture and color features in a image to tell road region apart from background \cite{QinH2010}. This is a tough task as the illumination and scene appearance may change dramatically and little useful prior information is available. Although, geometric priors like vanishing point detection \cite{Moghadam2011Fast} and horizon \cite{Almazan2016} has been introduced to improve road detection, the performance is still not so good in complex scenes. Since LiDAR is a basic sensor on autonomous land vehicle, many LiDAR based road detection algorithms have been discussed. Two widely used models are road geometry regression \cite{Chen2014,Hu2014,Asvadi2016,Hata2014} and grid cell classification. As a commonly used hypothesis in LiDAR based methods, road area are defined as a flat plane, thus, a lot of method detect road by finding out flat planes or analyzing height difference \cite{Chen2017}. In addition, 3D LiDAR road boundary detection is another popular way to detect road area \cite{Wijesoma2004}. Methods mentioned above rely so much on obvious height difference that they may result in poor performance in flat terrain. Fusion is a easy way to tackle the weakness of single sensor. Usually, point cloud from LiDAR are projected to image to correspond with camera and fusing strategy is various. Hu et al. in \cite{Hu2014} estimate road plane from point cloud and then screen out non-road points, the road points will be projected to image to fuse. Xiao et al. in \cite{Xiao2018} use an extended Condition Random Field(CRF) model to fuse image and LiDAR points. The CRF model make the best of information from both sensors instead of using them independently.

The recent raise of deep learning methods have made a big progression handling segmentation problems \cite{Simonyan2015} \cite{He2016} in complex scene. Vision based road detection methods use convolutional neural networks(CNN) to train a classifier for segmentation problems. Inspired by the great success of CNN, Long et al. in \cite{Long2015Fully} proposed fully convolutional network(FCN). In FCN, fully connected layers in VGG was replaced by convolutional layer and deconvolutional layers were added to up sampling feature maps. Road detection using CNNs and FCNs made tremendous progress \cite{Laddha2016,Caltagirone2017,Wang2018}. Gabriel et al. proposed a efficient deep model to speed up the road detection task and reached nearly real-time performance. In addition, Han et al. proposed a semi-supervised learning road detection using generative adversarial networks(GANs) to overcome insufficient training data in fully supervised learning schemes and achieved th state-of-the-art performance on KITTI ROAD benchmark \cite{Xiaofeng2018}.Deep learn method also has a lot of practice in LiDAR based road detection. Luca et al. in \cite{Luca2017} project LiDAR point to a top-view to create grid maps, with which a FCN can be used to perform fast pixel-wise road detection. This work is a top-performing algorithm on KITTI ROAD using LiDAR. Charles et al. design a novel type of neural network that directly consumes point clouds, which well respects the permutation invariance of points in the input, called PointNet \cite{Charles2017}. 

Our work is efficient and shows strong performance on par or even better than other works. It is a variation of Siamese network originally proposed by Y. Lecun\cite{Lecun1993}. Hinton in \cite{Hinton2010} used a Siamese architecture to make a binary classifier with two sets of faces. The key idea of this architecture is to learn the classifier by taking two different input that describe a single representation, which inspired the fusing mechanism in our work. There are three main contributions: firstly, a new network based on FCN-8s which embedded a siamese structure in encoder part to help with camera and LiDAR fusion; secondly, using sparse LiDAR image as input source instead of dense one, which can achieve similar performance with dramatically computation intensity reducing in processing pipeline; at last, evaluation of our work are performed in public dataset.

\section{Our Method and Insights}

Our method is a Siamese deep neural network based on FCN, which fuses data from LiDAR and image simultaneously. Though many methods on semantic segmentation have been proposed, the dominating  methods are developed based on FCN framework. Like many successful works, the proposed method follows the basic idea of encoder-decoder methodology in FCN. To fuse camera and LiDAR data, we redesign the encoder layers and a Siamese structure is embedded, thus, the network exists two branches. Our method separate the two input sources in the convolutional stage, and let them interact with each other in fusion layers. We believe our design will make better use of two different kinds of data. 
\begin{figure}[h]
	\centering
	\includegraphics[width=\textwidth]{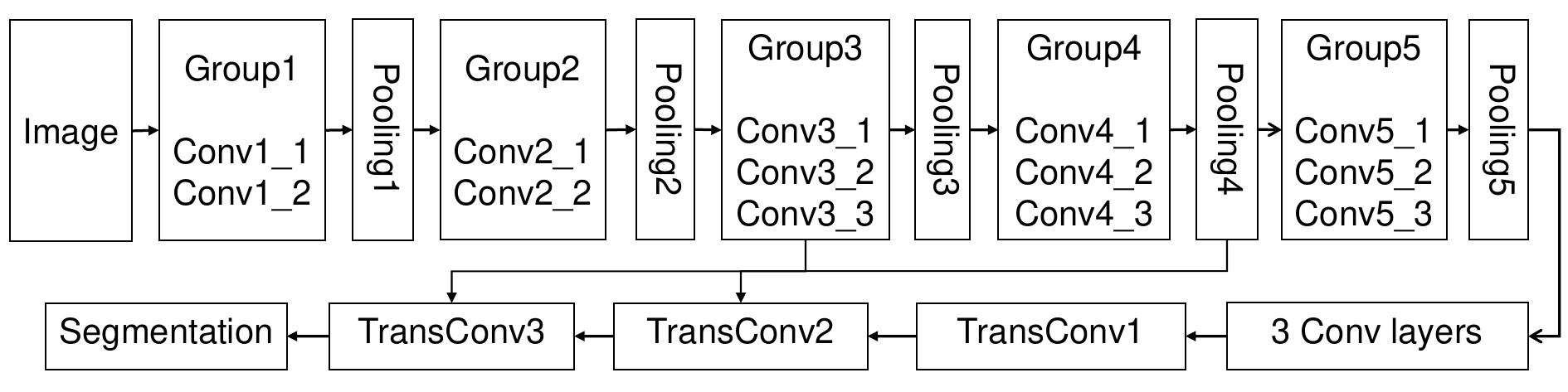}
	\caption{The architecture of FCN-8s. FCN-8s shares same convolutional structure with VGG-16. The differences lie in two aspects. Firstly, we replace 3 fully connected layers with 3 convolutional layers. Secondly, we attach 3 transpose convolutional layers to the sample feature maps.}
	\label{fig:FCN}
\end{figure}

\begin{figure}[h]
	\centering
	\includegraphics[width=\textwidth]{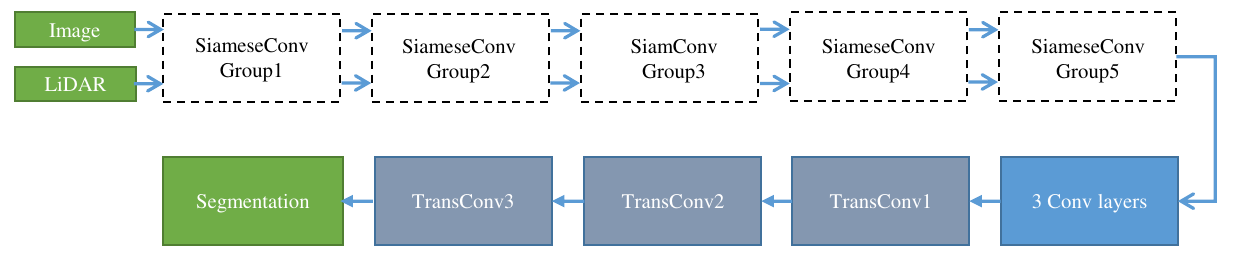}
	\caption{The proposed Siamese network architecture. Image and LiDAR data are fed to the network after been preprocessed, at the front of this pipeline are five consecutive Siamese convolutional groups. Like FCN-8s network, 3 convolutional layers and transpose convolutional layers take the result of front part mentioned above to generate segmentation result.}
	\label{fig:NetArh}
\end{figure}

There are five Siamese convolutional layer groups for data fusing in the proposed network in  Fig. \ref{fig:NetArh}. In each group, we extract features in two branches individually, and then concatenate them in a fusion layer. After that, one additional convolutional layer is used to perform a further fusion. Apart for the Siamese groups, we also set 5 pooling layers to down sample the feature maps and help our network learning in multiple scales. In each Siamese group, the output of each branch and the result of pooling layer are summed up separately, output as the input of its corresponding branch in the next Siamese group. After the fifth Siamese group, there are three convolutional layers in which the result is up sampled by 3 transpose convolutional layers to get dense road detection results. The following subsections will explain and discuss our designation details and insights.

\subsection{Sparse LiDAR Images}

\begin{figure}[h]
	\centering
	\includegraphics[width=\textwidth]{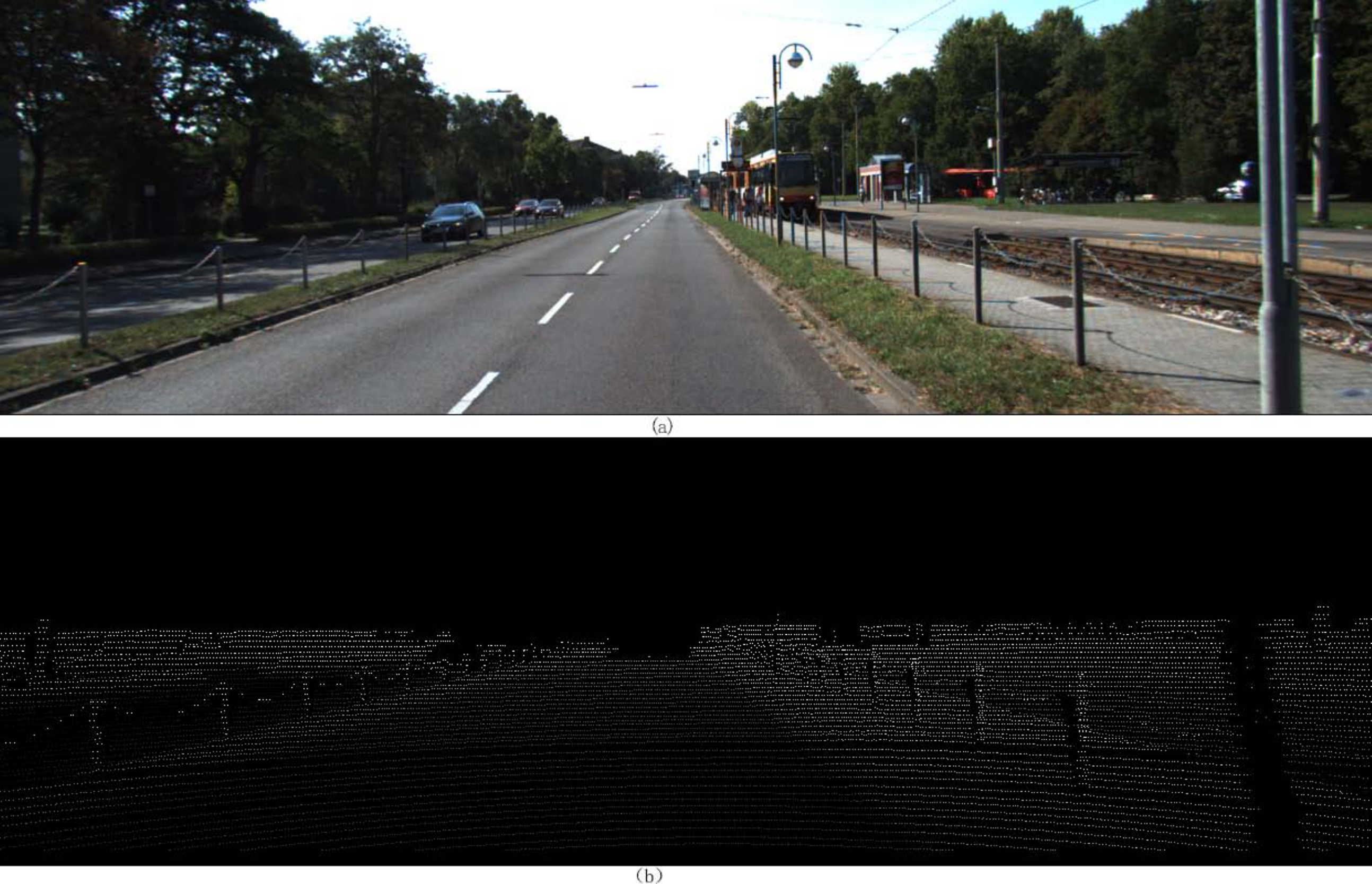}
	\caption{Project point clouds to get the LiDAR image. (a) is the RGB image and (b) is the LiDAR image. For demonstration, in the LiDAR image the higher the point is, the brighter the pixel will be.}
	\label{fig:proj}
\end{figure}
Our network has two heterogeneous input sources: RGB image and its corresponding LiDAR image. Unlike the RGB images, the LiDAR images are generated from projecting LiDAR point cloud to image plane. The projection need a set of calibration parameters between camera and LiDAR device. We assume the sensors are well calibrated in advance so that the projection matrix, including rotation and translation parameters, is already known. In practise, each frame of LiDAR data usually consist of 120,000 3D points with three location parameters and a intensity value. Only part of those points which can be projected to the RGB image plane will make sense in the following procedure. Let's denote the LiDAR point clouds as $\mathbf{p}_l=(x, y, z, 1)^\mathrm{T}$ and the projection results on RGB image plane as $\mathbf{p}_i=(u, v, 1)^\mathrm{T}$. The rotation matrix is $\mathbf{R} \in \mathbb{SO}(3)$ and the translation matrix is $\mathbf{t}\in \mathbb{R}^{3\times1}$. The intrinsic parameters of camera is $\mathbf{K}\in\mathbb{R}^{3\times3}$. Then the projection can be formulated as follows:
\begin{equation}
\centering
\mathbf{p}_i = \mathbf{K}\mathbf{T}\mathbf{p}_l \quad, \quad \mathbf{T} = {
\left[ \begin{array}{cc} 
    \mathbf{R} & \mathbf{t} \\
    \mathbf{0} & 1
\end{array}
\right]}
\end{equation}

Fig. \ref{fig:proj} shows one RGB image and its corresponding LiDAR image. To generate a LiDAR image, formula above is applied in the first place, hence, each 3D point finds its pixel location in LiDAR image. In our method, LiDAR images have 3 channels in which filled with 3 coordinate value of a 3D point. Pixels hit no 3D point are filled with zeros.

Compared with RGB image, the LiDAR image is much more sparse, nonetheless providing more geometrical information. By binging in LiDAR image as an input source, our network is capable of learning spacial features like height/depth or further complicated ones to improve road detection. Though there are some tricks in LiDAR image generation(such as normal vector extraction, hand-crafted edge detector , etc.), our method uses 3D coordination directly, since the convolutional neural network can extract features automatically in multiple scales. Due to the sparsity of LiDAR image, many other methods utilize up-sampling algorithms like MRF or joint bilateral filter to convert a sparse LiDAR image to dense one. However, up-sampling will dramatically increase the computation intensity. In our method, we consider the up-sampling step is not indispensable. We keep the sparsity of LiDAR image and make use of the encoder-decoder structure in FCN to bypass the manual up-sampling step with little performance loss. As a consequence, preprocessing latency in actual use is significantly reduced. To demonstrate our design, section 4 will provide a experiment result to evaluate.

\subsection{Fusion Strategies on FCN}
\label{strategy}

\begin{figure}[h]
	\centering
	\includegraphics[width=\textwidth]{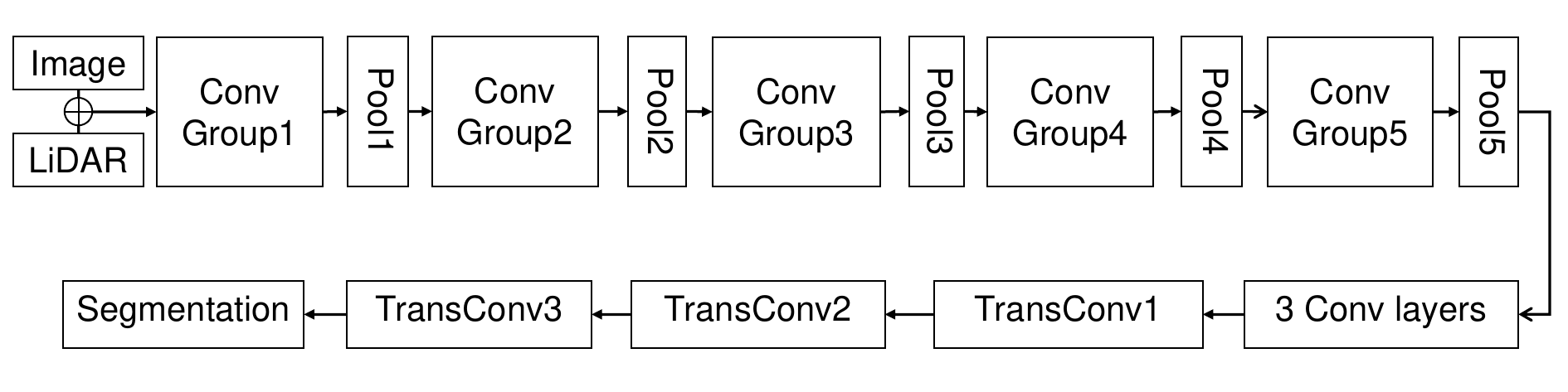}\\
	\includegraphics[width=\textwidth]{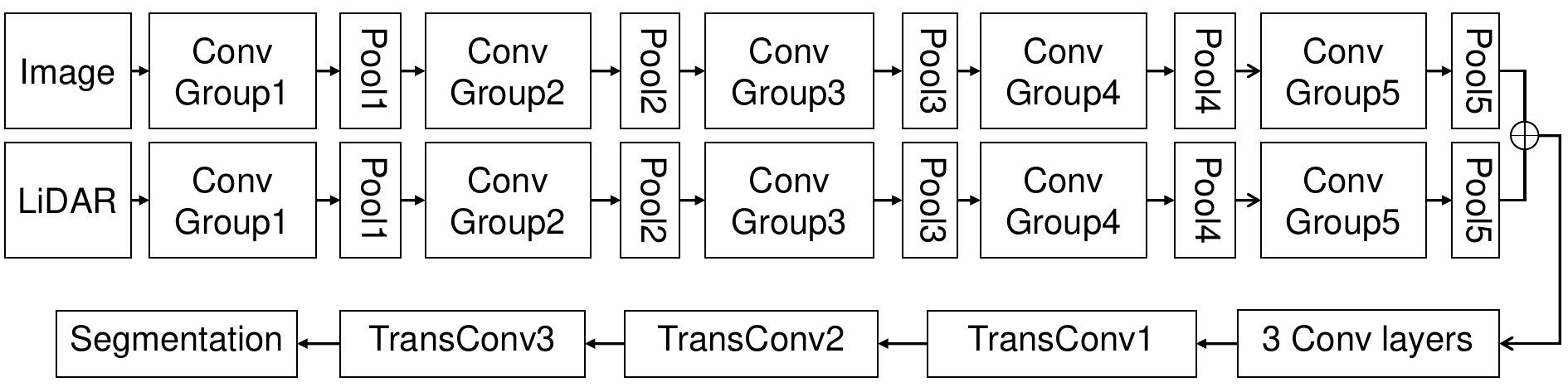}\\
	\caption{Different kinds of deep neural networks to fuse RGB image and LiDAR data. The first one fuses image and LiDAR data before feeding procedure. However, the second one feeds image and LiDAR data individually into two networks, information will be fused before transpose convolutional layers. For demonstration, they are all based on FCN-8s.}
	\label{fig:Fusion}
\end{figure}
\begin{figure}[h]
	\centering
	\includegraphics[width=0.8\textwidth]{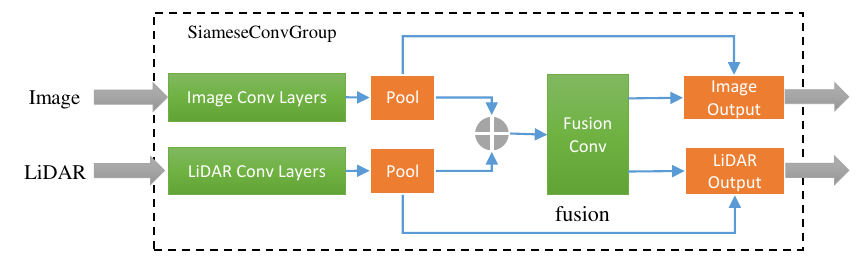}
	\caption{Architecture of Siamese Group.}
	\label{fig:siamesegroup}
\end{figure}

Since FCN is a series of network architecture for solving the semantic segmentation problems, we take FCN-8s as an example. Fig. \ref{fig:FCN} demonstrates the architecture of FCN-8s. There are two major ways to fuse RGB images and LiDAR point clouds under a deep neural network framework. For example, as shown in (a) in Fig. \ref{fig:Fusion}, they can fuse before feeding into the network, we call it early fusion. Besides, they can be fed into two individual branches separately and the results are fused before the transpose convolutional layers which is called late fusion, as (b) of Fig. \ref{fig:Fusion} shows. Both two ways have their strong points and weak points. Early fusion forces the network share parameters between color and LiDAR images, unfortunately, failed to balance two input sources. To fix the problem above, late fusion sets two separate convolution pipelines. The weakness is obvious in late fusion, due to the fact that interactions at each scale are banned, the result may be insufficient in multiple scales. To exploit these two kinds of data more effectively and make a much closer integration between their features in multiple scales, we propose a Siamese structure as shown in Fig. \ref{fig:siamesegroup} as our fusin strategy.

In this structure, color and LiDAR image can be input into to two convolutional branches respectively. After pooling and concatenation, a further convolutional operation follows to reduce the channel size. At last, pooling outputs will be added to this tensor respectively and output as the input of subsequent step. By replacing the original layers with Siamese structure, the network is capable of fusing image and LiDAR data in multiple scales as well as and balancing two input sources automatically.

\subsection{Siamese Fully Convolutional Network}
The standard FCN-8s network comprises five convolutional layer groups consist of several convolutional layers. We replace them with five groups of Siamese structure. Table \ref{tab:SiamConv} presents the details of our network. Both of the first two Siamese convolutional layer groups contain five convolutional layers: two layers denoted as "ImgConvX\_X" process RGB images, and two layers denoted as "LdaConvX\_X" process LiDAR images. Then, two feature maps from image and LiDAR branch are concatenated, thus the channel numbers are doubled. 

However, large feature dimensions will lead to extra costs of resources. The following convolutional layer in a group called "FuseConvX" is used to reduce the dimensions as well as fuse data at the same time. Compared with the first two groups, the subsequent Siamese convolutional layer groups share the similar structure but different more convolutional steps. Unlike first two Siamese groups, the rest three Siamese groups set the number of convolutional layers for each branch to three, just like the VGG-16 and the FCN does. In the decoder part, which has three transpose convolutional layers, the feature maps will be up sampled after the first and second  transpose convolutional layer. Eventually, the output of the third transpose convolutional layer and the original RGB image will equal in size. Also, two skip layers like the FCN-8s are kept in the proposed newtwork.

\begin{table}
	\centering
	\caption{Details of Siamese Fully Convolutional Network}
	\label{tab:SiamConv}
	\begin{tabular}{ccccc}
		\hline\noalign{\smallskip}
		Group Name & Layer Name & Kernel Size & Stride & Feature Map Size\\
		\noalign{\smallskip}\hline\noalign{\smallskip}
		\multirow{5}{*}{SiamGroup1}
		~ & ImgConv1\_1 & (3,3,3,64) & 1 & (375,1242)\\
		~ & ImgConv1\_2 & (3,3,64,64) & 1 & (375,1242)\\
		~ & LdaConv1\_1 & (3,3,3,64) & 1 & (375,1242)\\
		~ & LdaConv1\_2 & (3,3,64,64) & 1 & (375,1242)\\
		~ & FuseConv1 & (3,3,128,64) & 1 & (375,1242)\\
		\hline
		\multirow{5}{*}{SiamGroup2}
		~ & ImgConv2\_1 & (3,3,64,128) & 1 & (188,621)\\
		~ & ImgConv2\_2 & (3,3,128,128) & 1 & (188,621)\\
		~ & LdaConv2\_1 & (3,3,64,128) & 1 & (188,621)\\
		~ & LdaConv2\_2 & (3,3,128,128) & 1 & (188,621)\\
		~ & FuseConv2 & (3,3,256,128) & 1 & (188,621)\\
		\hline
		\multirow{7}{*}{SiamGroup3}
		~ & ImgConv3\_1 & (3,3,128,256) & 1 & (94,311)\\
		~ & ImgConv3\_2 & (3,3,256,256) & 1 & (94,311)\\
		~ & ImgConv3\_3 & (3,3,256,256) & 1 & (94,311)\\
		~ & LdaConv3\_1 & (3,3,128,256) & 1 & (94,311)\\
		~ & LdaConv3\_2 & (3,3,256,256) & 1 & (94,311)\\
		~ & LdaConv3\_3 & (3,3,256,256) & 1 & (94,311)\\
		~ & FuseConv3 & (3,3,512,256) & 1 & (94,311)\\
		\hline
		\multirow{7}{*}{SiamGroup4}
		~ & ImgConv4\_1 & (3,3,256,512) & 1 & (47,156)\\
		~ & ImgConv4\_2 & (3,3,512,512) & 1 & (47,156)\\
		~ & ImgConv4\_3 & (3,3,512,512) & 1 & (47,156)\\
		~ & LdaConv4\_1 & (3,3,256,512) & 1 & (47,156)\\
		~ & LdaConv4\_2 & (3,3,512,512) & 1 & (47,156)\\
		~ & LdaConv4\_3 & (3,3,512,512) & 1 & (47,156)\\
		~ & FuseConv4 & (3,3,1024,512) & 1 & (47,156)\\
		\hline
		\multirow{7}{*}{SiamGroup5}
		~ & ImgConv5\_1 & (3,3,512,512) & 1 & (24,78)\\
		~ & ImgConv5\_2 & (3,3,512,512) & 1 & (24,78)\\
		~ & ImgConv5\_3 & (3,3,512,512) & 1 & (24,78)\\
		~ & LdaConv5\_1 & (3,3,512,512) & 1 & (24,78)\\
		~ & LdaConv5\_2 & (3,3,512,512) & 1 & (24,78)\\
		~ & LdaConv5\_3 & (3,3,512,512) & 1 & (24,78)\\
		~ & FuseConv5 & (3,3,1024,512) & 1 & (24,78)\\
		\hline
		\multirow{3}{*}{TransGroup}
		~ & TransConv1 & (4,4,2,512) & (24,78) & (47,156)\\
		~ & TransConv2 & (4,4,512,256) & (47,156) & (94,311)\\
		~ & TransConv3 & (16,16,256,2) & (94,311) & (375,1242)\\
		\noalign{\smallskip}\hline
	\end{tabular}
\end{table}

\section{Experiments}

To verify the performance of our method, we carry out experiments on KITTI ROAD dataset \cite{KITTI}. The experimental details and results are introduced in the following subsections.

\subsection{Dataset and Experimental Setting}

This dataset contains 579 frames of color images, along with their corresponding LiDAR point clouds. The data are collected by a moving vehicle in urban environment with an accurate calibration. There are many kinds of backgrounds such as pedestrians, trees, buildings, fences and vehicles. Also on the road surfaces there are various sizes of shadows and some lane lines are not very clear. The resolution of color images is 375$\times$1242 pixels. 289 frames are used as training data and 290 frames are for testing. The data are classified into 3 categories: UM (urban marked), UMM (urban multiple marked lanes) and UU (urban unmarked). The numbers of frames in training dataset of each category are 95, 96 and 98, while the numbers of them in testing dataset are 96, 94, 100.

Our networks are implemented using Tensorflow framework and trained on an NVIDIA TITAN X GPU with 12GB on board memory. The resolution of input RGB/LiDAR images in KITTI dataset is 375$\times$1242 in pixels. The batch size is set to one during training, which means, we feed only 1 RGB image and its corresponding LiDAR image into the network for each training step. The initial learning rate is $10e^{-6}$, and will be successively halved every 5000 iterations. The iteration number is 60,000. Our network is trained from scratch since there are no similar dataset.

\subsection{Performances of different fusion architectures}
To verify the road detection performance of different fusion strategies, we split the training dataset into two parts: the training data set has 240 frames of data and the rests are used as the validation data. Data in both parts are randomly selected from all the three categories. After that, the three fusion strategies mentioned above are tested on the training dataset. The early fusion and late fusion is described in Section \ref{strategy}. The F-measure, the precision, the recall and the accuracy are listed in Table \ref{tab:PerFusion} to evaluate the performances. 

\begin{table}[h]
	\centering
	\caption{Performances of different fusion strategies}
	\label{tab:PerFusion}
	\begin{tabular}{ccccc}
		\hline\noalign{\smallskip}
		Fusion Strategy & MaxF & PRE & REC & Accuracy\\
		\noalign{\smallskip}\hline\noalign{\smallskip}
		Early Fusion & 89.68\% & 90.02\% & 88.71\% & 95.59\% \\
		Late Fusion & 90.87\% & 91.13\% & 90.79\% & 96.88\%\\
		Siamese structure & \textbf{91.40\%} & \textbf{91.21\%} & \textbf{91.60\%} & \textbf{97.02\%} \\
		\noalign{\smallskip}\hline
	\end{tabular}
\end{table}
According to Table \ref{tab:PerFusion}, our Siamese structure outperforms others since we fuse features from RGB images and LiDAR images in every scales. The early fusion lagging behind in performance because two information are integrated in early stage. Late fusion perform better because features are extracted in parallel branches, thus more usefully features can be retained. Our Siamese structure uses a gradual fusion strategy in which fusion procedure is added after a short late fusion in every group, which improved data fusion in multiple scales.

\subsection{Performances of Siamese network using sparse and dense LiDAR images}

\begin{figure}[h]
	\centering
	\includegraphics[width=\textwidth, height=0.7\textwidth]{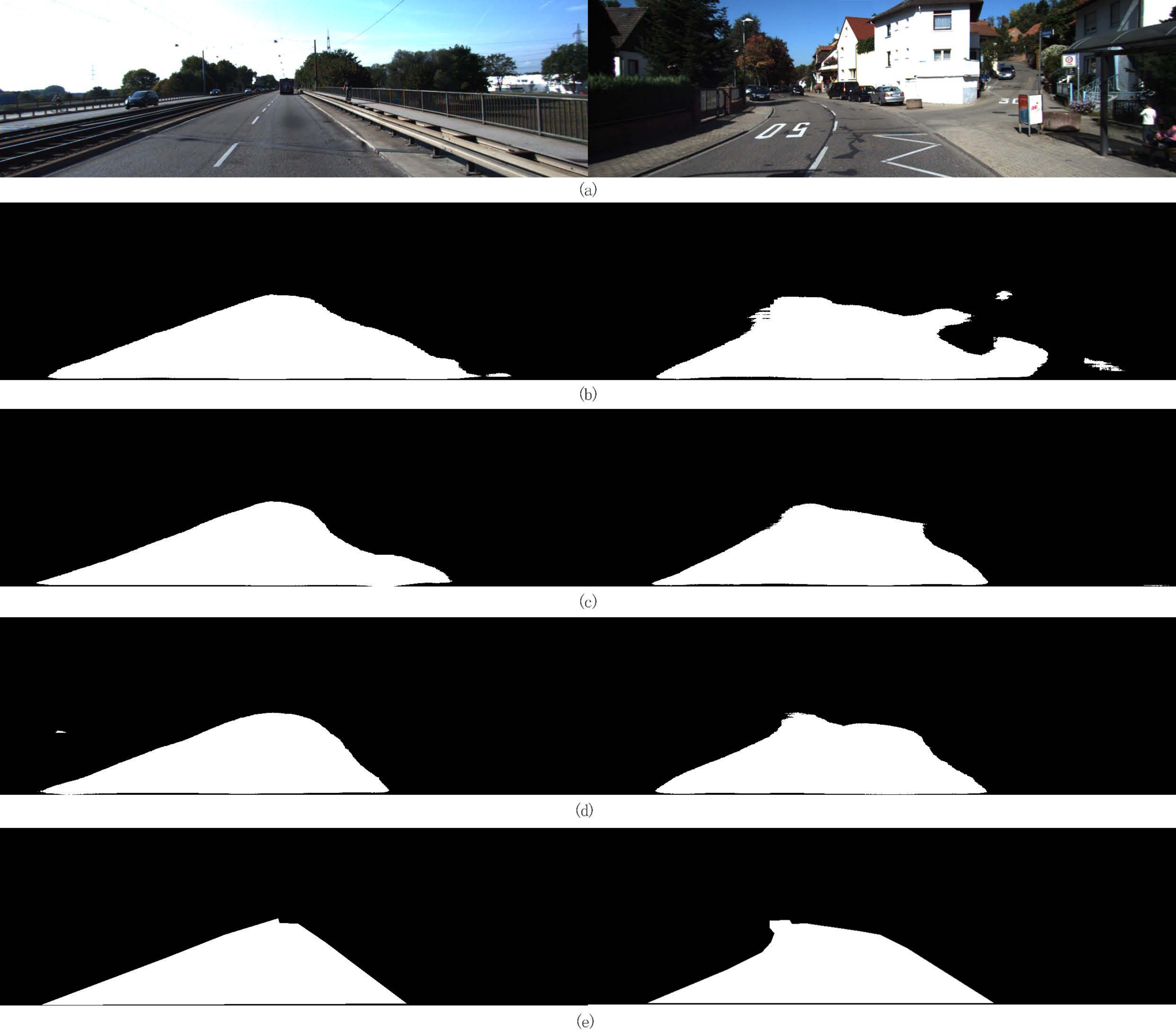}
	\caption{The road detection results of different methods. (a) are two RGB images, and from (b) to (d) are the results of FCN-8s with RGB images, Siamese-FCN with dense LiDAR images and Siamese-FCN with sparse LiDAR images. (e) are the ground truths.}
	\label{fig:FCNandSiamFCN}
\end{figure}

To verify the performance of our proposed network using sparse and dense LiDAR images, we train our network twice on the above training dataset and then test them on validation dataset. The Dense LiDAR image is generated by MRF. The results are shown in Table \ref{tab:PerOnVal}. The data demonstrates that our Siamese network can improve the road detection performance, even using sparse and dense LiDAR images.
\begin{table}[h]
	\caption{Performances of using sparse and dense LiDAR images}
	\label{tab:PerOnVal}
	\begin{tabular}{cccccc}
		\hline\noalign{\smallskip}
		Input type & MaxF & PRE & REC & Accuracy & Time \\
		\noalign{\smallskip}\hline\noalign{\smallskip}
		Original FCN-8s (RGB only) & 88.31\% & 89.46\% & 87.19\% & 96.01\% & 0.07s\\
		Siamese-FCN (dense LiDAR) & 91.86\% & 92.73\% & 91.02\% & 97.21\% & 2.25s\\
		Siamese-FCN (sparse LiDAR) & 91.40\% & 91.21\% & 91.60\% & 97.02\% & 0.18s\\
		\noalign{\smallskip}\hline
	\end{tabular}
\end{table}
We show two road detection results of these three networks in Fig. \ref{fig:FCNandSiamFCN}. In these two scenes, the result of FCN-8s trained on RGB images have much more false positives than others, since the sidewalk area and road area in other training images are exactly similar. Too much resemblance in color and subtle difference in texture makes it quite hard too distinguish the sidewalk from road in many pure image based methods. However, with the help of LiDAR images, these errors can be restrained as the 3D structure of sidewalk areas and roads are very different. By using dense LiDAR images, the result is improved, but due to the up sampling processing under the MRF framework, the inference time increases significantly with a slight performance improvement. After a trade off between time and accuracy, we only use the sparse LiDAR images in our method.

\subsection{Road Detection Performance}

To evaluate road detection performance, we trained our network on the whole training dataset and then upload the road detection results on testing dataset to the KITTI benchmark server. Our method detects road regions in all tree categories and dose not distinguish different lanes in UMM. A set of prescribed metrics in bird's eye view(BEV) images are used for evaluation, namely maximum F1-measure (MaxF), average precision (AP), precision (PRE), recall (REC), false positive rate (FPR) and false negative rate (FNR).
\begin{table}[h]
	\caption{Performances on different categories of KITTI ROAD benchmark}
	\label{tab:PerKITTI1}
	\centering
	\begin{tabular}{ccccccc}
		\hline
		Benchmark	& MaxF	& AP	& PRE	& REC	&  FPR	& FNR\\
		\hline
		UM ROAD		& 91.03\%	& 84.64\% 	& 89.98\% 	& 92.11\% 	& 4.67\% 	& 7.89\%\\
		UMM ROAD    & 93.68\% 	& 89.74\%	& 93.48\% 	& 93.87\% 	& 7.20\% 	& 6.13\%\\
		UU ROAD		& 88.02\% 	& 75.58\% 	& 86.91\% 	& 89.16\% 	& 4.37\% 	& 10.84\%\\
		URBAN ROAD	& 91.51\% 	& 85.79\% 	& 90.82\% 	& 92.21\% 	& 5.13\% 	& 7.79\%\\
		\hline
	\end{tabular}
\end{table}
Table \ref{tab:PerKITTI1} shows the results of our method in 3 categories and urban dataset. Our method performs better on UM ROAD and UMM ROAD than UU ROAD. In these two categories there are usually many curbs or other road boundaries that usually provide many spacial features obviously separate road and non-road areas in 3D coordinate system in LiDAR branch. Therefore, with the help of LiDAR images, our Siamese network can learn from that appearance and outperform others after fuse.

\begin{figure}[h]
	\centering
	\includegraphics[width=0.8\textwidth, height=0.5\textwidth]{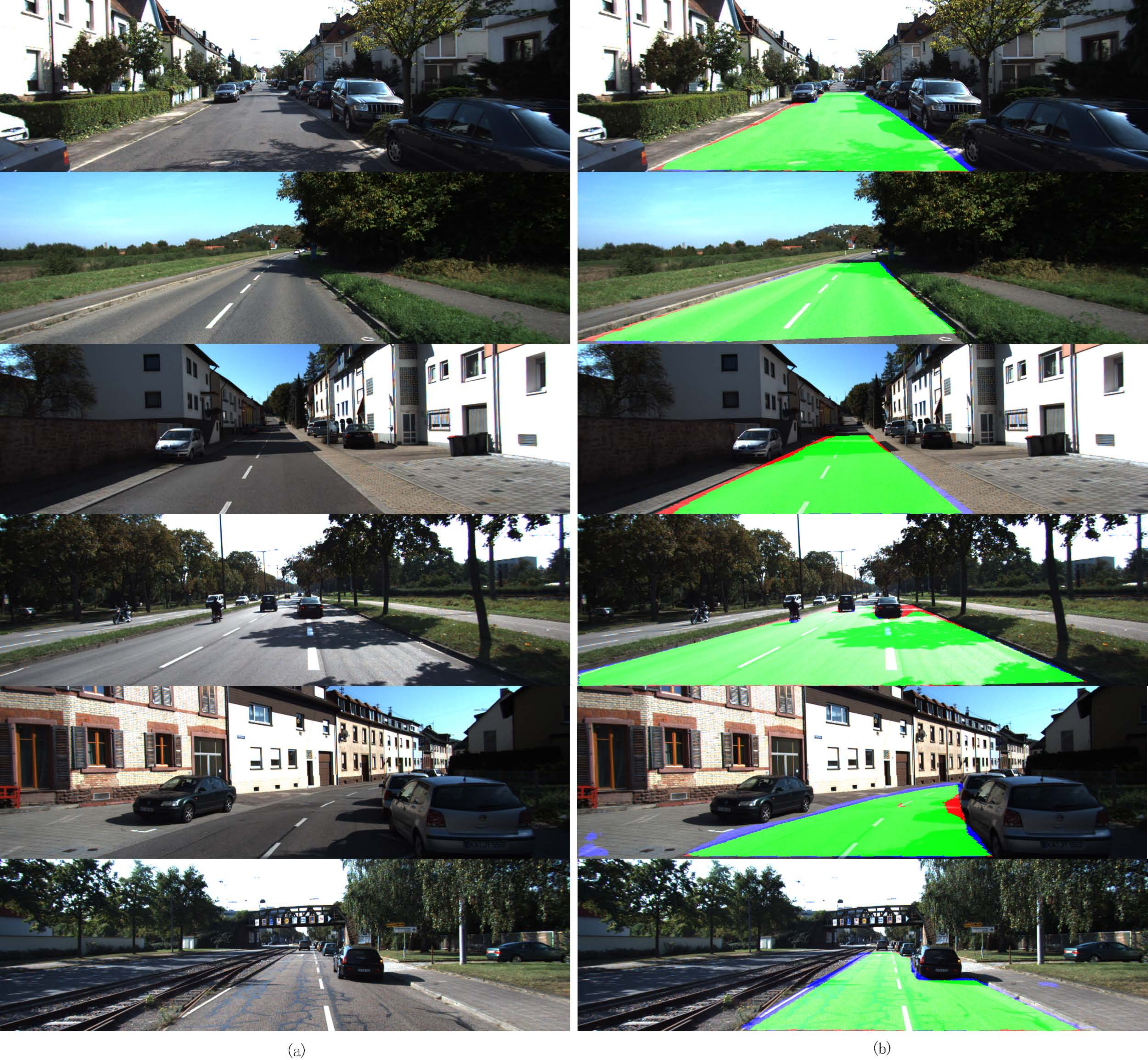}
	\caption{The road detection results. (a) are the RGB images and (b) are the road detection results.}
	\label{fig:res}
\end{figure}

\begin{table}
	\centering
	\renewcommand{\arraystretch}{1.2}
	\caption{Performances of different methods on KITTI ROAD benchmark}
	\label{tab:PerKITTI2}
	\centering
	\begin{tabular}{ccccccc}
		\hline
		Method	& MaxF	& AP	& PRE	& REC	& FPR	& FNR\\
		\hline
		Multi-task CNN\cite{mtCNN}	& 86.81\%	& 82.15\%	& 78.26\%	& 97.47\%	& 14.92\%	& 2.53\%\\
		FCN-LC\cite{FCN-LC}			& 90.79\% 	& 85.83\% 	& 90.87\% 	& 90.72\% 	& 5.02\% 	& 9.28\%\\
		LidarHisto\cite{Chen2017}	& 90.67\% 	& 84.79\% 	& 93.06\% 	& 88.41\% 	& 3.63\% 	& 11.59\%\\
		MixedCRF\cite{Han2016}		& 90.59\% 	& 84.24\% 	& 89.11\% 	& 92.13\% 	& 6.20\% 	& 7.87\%\\
		HybridCRF\cite{HybridCRF}	& 90.99\%	& 85.26\%	& 90.65\%	& 91.33\%	& 4.29\%	& 8.67\%\\
		FusedCRF\cite{FusedCRF}		& 89.55\%	& 80.00\%	& 84.87\%	& 94.78\%	& 7.70\%	& 5.22\%\\
		Our method					& 91.51\% 	& 85.79\% 	& 90.82\% 	& 92.21\% 	& 5.13\% 	& 7.79\%\\
		\hline
	\end{tabular}
\end{table}
The results of some other methods and ours on the whole KITTI ROAD dataset are shown in Table \ref{tab:PerKITTI2}. They are Multi-task CNN, FCN-LC, LidarHisto, MixedCRF, HybridCRF and FusedCRF. The first 2 methods are deep learning based methods while others are mainly CRF-based methods. Our method outperforms other methods, even it's trained from scratch and we do not apply any kinds of post-processing such as CRF. Though we use sparse LiDAR image as our input, the result is still good in precision and recall. The precision is not very good since image branch dragged the performance compared with pure LiDAR based method. Fig. \ref{fig:res} shows our final results on the KITTI ROAD benchmark in the perspective images. In the images, red areas denote false negatives, blue areas correspond to false positives and green area represent true positives.

\section{Conclusion and Future Works}

In this paper, we proposed a Siamese deep neural network to detect road via RGB image and LiDAR point clouds. The network has two branches based on FCN-8s. We project the LiDAR point clouds onto the images to generate sparse LiDAR images, in which some pixels have their 3D location values. The RGB images and LiDAR images are fed into each branch separately but fused in each scale to improve segmentation performance. We verify our method on KITTI ROAD dataset and the results show its effectiveness. However, there are still some challenging issues in this problem, For example the road edge areas are not classified very well and available dataset is too small. We will continue working on it in our future works.

\section{Acknowledgments}

This research was supported by the Major Special Project of Core Electronic Devices, High-end Generic Chips and Basic Software(Grant No. 2015ZX01041101), National Defense Pre-research Foundation(Grant No.41412010101) and the China Postdoctoral Science Foundation (Grant No. 2016M600433).

\section{Notes}
In this section, we list some details for the notation and indicators mentioned above. In following equations, $T$ is short for TRUE, $F$ is short for FALSE, $P$ is short for POSITIVE and $N$ is short for NEGATIVE. The definition is shown as follows:
\begin{eqnarray*}
&& PRE(precision) = TP/(TP+FP) \\
&& REC(recall) = TP/(TP+FN) \\
&& MaxF(F\textrm{-}measure) = 2\times PRE \times REC/(PRE+REC) \\
&& FPR = FP/(TN+FP) \\
&& FNR = FN/(FN+FP) \\
&& Accuracy = (TP+TN)/(TP+TN+FP+FN)
\end{eqnarray*}

\end{document}